\documentclass[sigconf]{acmart}
\renewcommand\footnotetextcopyrightpermission[1]{} 
\AtBeginDocument{%
  }

\setcopyright{acmlicensed}
\copyrightyear{2018}
\acmYear{2018}
\acmDOI{XXXXXXX.XXXXXXX}

\acmConference[Conference acronym 'XX]{Make sure to enter the correct
  conference title from your rights confirmation emai}{June 03--05,
  2018}{Woodstock, NY}
\acmISBN{978-1-4503-XXXX-X/18/06}

\begin{document}

\title{ExVideo: Extending Video Diffusion Models via Parameter-Efficient Post-Tuning}


\author{Zhongjie Duan}
\affiliation{
  \institution{East China Normal University}
  \city{Shanghai}
  \country{China}}
\email{zjduan@stu.ecnu.edu.cn}

\author{Wenmeng Zhou}
\affiliation{
  \institution{Alibaba Group}
  \city{Hangzhou}
  \country{China}}
\email{wenmeng.zwm@alibaba-inc.com}

\author{Cen Chen}
\affiliation{
  \institution{East China Normal University}
  \city{Shanghai}
  \country{China}}
\email{cenchen@dase.ecnu.edu.cn}

\author{Yaliang Li}
\affiliation{
  \institution{Alibaba Group}
  \city{Hangzhou}
  \country{China}}
\email{yaliang.li@alibaba-inc.com}

\author{Weining Qian}
\affiliation{
  \institution{East China Normal University}
  \city{Shanghai}
  \country{China}}
\email{wnqian@dase.ecnu.edu.cn}

\renewcommand{\shortauthors}{Duan et al.}

\begin{abstract}
  Recently, advancements in video synthesis have attracted significant attention. Video synthesis models such as AnimateDiff and Stable Video Diffusion have demonstrated the practical applicability of diffusion models in creating dynamic visual content. The emergence of SORA has further spotlighted the potential of video generation technologies. Nonetheless, the extension of video lengths has been constrained by the limitations in computational resources. Most existing video synthesis models can only generate short video clips. In this paper, we propose a novel post-tuning methodology for video synthesis models, called ExVideo. This approach is designed to enhance the capability of current video synthesis models, allowing them to produce content over extended temporal durations while incurring lower training expenditures. In particular, we design extension strategies across common temporal model architectures respectively, including 3D convolution, temporal attention, and positional embedding. To evaluate the efficacy of our proposed post-tuning approach, we conduct extension training on the Stable Video Diffusion model. Our approach augments the model's capacity to generate up to $5\times$ its original number of frames, requiring only 1.5k GPU hours of training on a dataset comprising 40k videos. Importantly, the substantial increase in video length doesn't compromise the model's innate generalization capabilities, and the model showcases its advantages in generating videos of diverse styles and resolutions. We will release the source code and the enhanced model publicly\footnote{Project page: \url{https://ecnu-cilab.github.io/ExVideoProjectPage}\\ Github repo: \url{https://github.com/modelscope/DiffSynth-Studio}}.
\end{abstract}





\maketitle

\section{Introduction}


Over recent years, diffusion models \cite{sohl2015deep, ho2020denoising} have achieved outstanding results in image synthesis, significantly surpassing previous GANs \cite{dhariwal2021diffusion}. These achievements have subsequently fostered a burgeoning interest in the adaptation of diffusion models for video synthesis. Models such as Stable Video Diffusion \cite{blattmann2023stable}, AnimateDiff \cite{guo2023animatediff}, and VideoCrafter \cite{chen2023videocrafter1} epitomize this research trajectory, showcasing the ability to produce frames that are not only coherent but also of high visual quality. These achievements underscore the practicality and potential of employing diffusion models in the field of video synthesis. With the groundbreaking results of SORA \cite{liu2024sora} reported at the beginning of 2024, the research direction of video synthesis has once again attracted widespread attention.

Although current video synthesis models are capable of producing video clips of satisfactory quality, there remains a considerable gap in their ability to extend the duration of the videos generated. To generate longer videos, current methodologies can be categorized into three types. 1) Training with long video datasets \cite{chen2024panda, wang2023internvid, bain2021frozen}: Given the computational constraints, present video generation models are predominantly trained on short video clips. The use of longer videos for training would result in prohibitively escalated costs. 2) Generating videos in a streaming \cite{kodaira2023streamdiffusion} or sliding window \cite{duan2024diffutoon} manner. Without further training, we can process the video by dividing it into multiple segments. However, this approach leads to lower video coherence. In addition, existing video generation models lack the capability for long-term video understanding, hence the accumulation of errors becomes inevitable. As a result, during the generation of long videos, the visual quality is prone to deterioration, manifesting as a breakdown in the imagery. 3) Frame Interpolation \cite{huang2022real, wu2024perception}: Video frame interpolation models offer a method to augment the frame count of generated videos. However, this approach is inadequate for extending the narrative timeframe of the video. While it increases the number of frames, retaining the original frame rate would give the video an unnatural slow-motion effect, failing to prolong the narrative span of the video content. These outlined challenges underscore the necessity for innovative solutions capable of overcoming the existing hurdles associated with video duration extension, without compromising video quality or coherence.

Recent breakthroughs in the development of LLMs (large language models) \cite{xiong2023effective, xiao2023efficient, chen2023longlora} have inspired us. Notably, LLMs, despite being trained on fixed-length data, exhibit remarkable proficiency in understanding contexts of variable lengths. This flexibility is further enhanced through the integration of supplementary components and the application of lightweight training procedures, enabling the processing of exceptionally lengthy texts. Such innovations have motivated us to explore analogous methodologies within video synthesis models. In this paper, we introduce a novel post-tuning strategy, called ExVideo, specifically designed to empower existing video synthesis models to produce extended-duration videos within the constraints of limited computational resources. We have designed an extension structure for mainstream video synthesis model architectures. This framework incorporates adapter components, meticulously engineered to maintain the intrinsic generalization capabilities of the base model. Through post-tuning, we enhance the temporal modules of the model, thereby facilitating the processing of content across longer temporal spans.

In theory, ExVideo is designed to be compatible with the majority of existing video synthesis models. To empirically validate the efficacy of our post-tuning methodology, we applied it to the Stable Video Diffusion model \cite{blattmann2023stable}, a popular open-source image-to-video model. Through ExVideo, we can extend the original frame production capacity from a limit of 25 frames to 128 frames. Importantly, this expansion was achieved without compromising the model's distinguished generative capabilities. Additionally, the enhanced model exhibits the versatility to be seamlessly integrated with text-to-image models \cite{podell2023sdxl, li2024hunyuan, chen2023pixart}. This synergistic amalgamation establishes robust and versatile pipelines. This adaptability underscores the potential of our post-training technique, the source code and the extended model will be released publicly.

In summary, the contributions of this paper include:
\begin{itemize}
    \item We present ExVideo, a post-tuning technique for video synthesis models that can extend the temporal scale of existing video synthesis models to facilitate the generation of long videos.
    \item We have trained an extended video generation model based on Stable Video Diffusion, capable of generating coherent videos of up to 128 frames while preserving the generative capabilities of the original model.
\end{itemize}

\section{Related Work}

\subsection{Diffusion Models}

Diffusion models \cite{sohl2015deep, ho2020denoising} are a category of generative models that model the generation process as a Markov random process. Unlike GANs \cite{goodfellow2014generative}, diffusion models do not require adversarial training, hence their training process is more stable. Moreover, through an iterative generation process, diffusion models are capable of producing images of exceptionally high quality. In recent years, image synthesis models based on diffusion, including Pixart \cite{chen2023pixart}, Imagen \cite{saharia2022photorealistic}, Hunyuan DiT \cite{li2024hunyuan}, and the Stable Diffusion series \cite{rombach2022high, podell2023sdxl, kang2024distilling}, have achieved impressive success. Diffusion models have given rise to a vast open-source technology ecosystem. Technologies such as LoRA \cite{hu2021lora}, ControlNet \cite{zhang2023adding}, DreamBooth \cite{ruiz2023dreambooth}, Textual Inversion \cite{gal2022image}, and IP-Adapter \cite{ye2023ip} have endowed the generation process of diffusion models with a high degree of controllability, thereby meeting the needs of various application scenarios.

\subsection{Video Synthesis}

Given the remarkable success of diffusion models in image synthesis, video synthesis approaches based on diffusion have also been proposed in recent years. For example, by adding motion modules to the UNet model \cite{ronneberger2015u} in Stable Diffusion \cite{rombach2022high}, AnimateDiff \cite{guo2023animatediff} transfers the capabilities of image synthesis to video synthesis. Stable Video Diffusion \cite{blattmann2023stable} is an image-to-video model architecture and can synthesize video clips after end-to-end video synthesis training. Unlike image synthesis models, video synthesis models require substantial computational resources since the model needs to process multiple frames simultaneously. As a result, most current video generation models can only produce very short video clips. For instance, AnimateDiff can generate up to 32 frames, while Stable Video Diffusion can generate a maximum of 25 frames. This limitation prompts us to explore how to construct video synthesis models over longer temporal scales.

\subsection{Extending Generative Models}

Although the existing diffusion models are trained with a fixed scale, such as Stable Diffusion being trained at a fixed resolution of $512\times 512$, some approaches can extend them to larger scales. For instance, in image synthesis, approaches like Mixture of Diffusers \cite{jimenez2023mixture}, MultiDiffusion \cite{bar2023multidiffusion}, and ScaleCrafter \cite{he2023scalecrafter} can increase the resolution of generated images by altering the inference process of the UNet model in Stable Diffusion. Similar techniques have also emerged in large language models. With the help of positional encoding technologies such as RoPE \cite{su2024roformer} and ALiBi \cite{press2021train}, large language models can extrapolate to longer text processing tasks under the premise of training with limited-length texts. Post-tuning can further help language models achieve super-long text comprehension and generation \cite{xiong2023effective, chen2023longlora}. These research findings have inspired us, prompting us to explore the extension of video synthesis models. We aim to endow existing video synthesis models with the capability to generate longer videos.

\section{Methodology}

\begin{figure*}
  \includegraphics[width=1.0\linewidth]{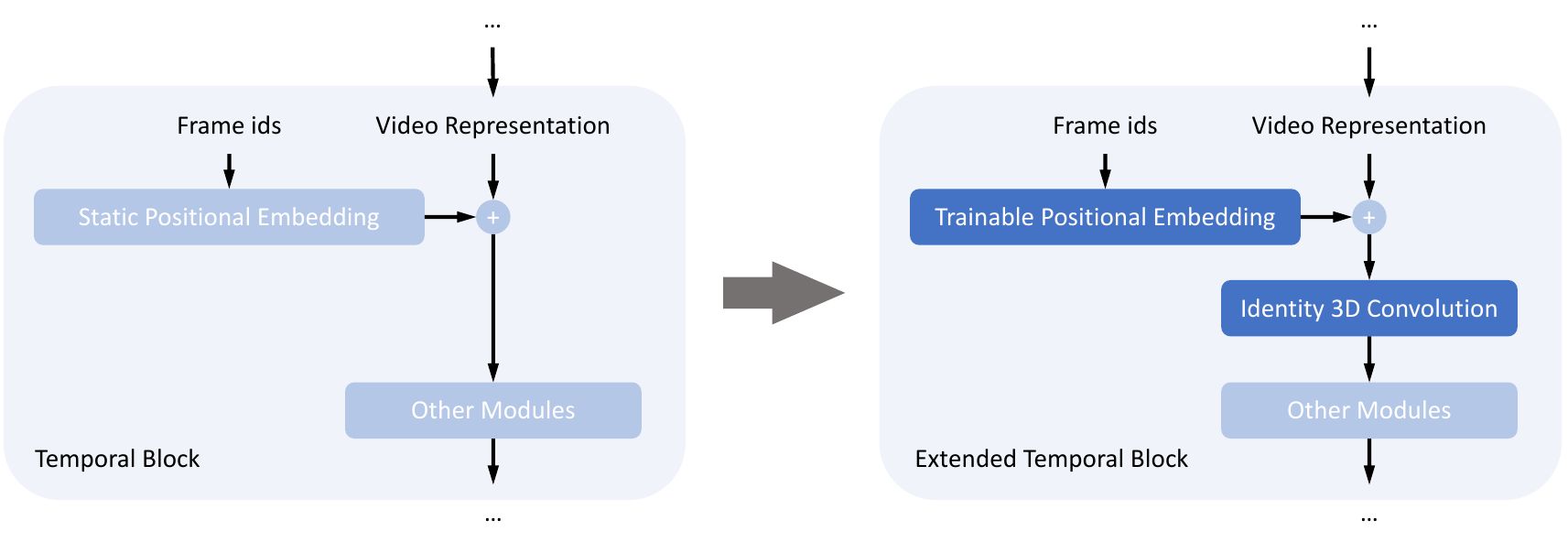}
  \caption{The architecture of extended temporal blocks in Stable Video Diffusion. We replace the static positional embedding with trainable positional embedding and add an adaptive identity 3D convolution layer to learn long-term video features. The modifications are adaptive, preserving the original generalization abilities of the pre-trained model. All parameters outside the temporal block are fixed while training for lower memory usage.}
  \label{figure:architecture}
\end{figure*}

In this section, we first review the architectures of mainstream video diffusion models, then discuss how we extend the temporal modules for long video synthesis, and finally introduce the post-tuning strategy.

\subsection{Preliminaries}

The huge demands for computational resources in training video synthesis models lead to a prevalent practice of adapting existing image synthesis models for video generation. This adaptation is typically achieved by incorporating temporal modules into the model to enable the synthesis of dynamic content. We provide a comprehensive overview of temporal module architectures, which can be categorized as follows:
\begin{itemize}
    \item \textbf{3D convolution} \cite{li2021survey}: Convolution layers form the foundational blocks in computer vision. 2D convolution layers have been employed in the UNet \cite{ronneberger2015u} architecture, which is widely used in diffusion models. By extending 2D convolutions into the third dimension, these layers are seamlessly adapted for use in video synthesis models. Research indicates that convolution layers in diffusion models exhibit a high degree of adaptability across various resolutions \cite{bar2023multidiffusion}, which is a testament to their capacity for generalization.
    \item \textbf{Temporal attention} \cite{vaswani2017attention}: In image synthesis, the importance of attention mechanisms is underscored by their contribution to the generation of images with remarkable fidelity, as evidenced by the ablation studies in latent diffusion \cite{rombach2022high}. Transferring spatial attention mechanisms to the video domain raises concerns regarding computational efficiency due to the quadratic time complexity of the attention operators. To circumvent this computational bottleneck, advanced video synthesis models typically adopt additional temporal attention layers \cite{guo2023animatediff, blattmann2023stable} that optimize efficiency by curtailing the volume of embeddings processed by each attention operator.
    \item \textbf{Positional embedding} \cite{su2024roformer}: The native attention layers cannot model the positional information in videos. Therefore, video synthesis models typically incorporate positional embeddings to enrich the embedding space with positional information. Positional embeddings can be instantiated through diverse methodologies. For example, AnimateDiff \cite{guo2023animatediff} opts for learnable parameters to establish positional embeddings, whereas Stable Video Diffusion \cite{blattmann2023stable} utilizes trigonometric functions to generate static positional embeddings.
\end{itemize}

\subsection{Extending Temporal Modules}

Most video synthesis models are pre-trained on videos comprising only a constrained number of frames due to limited computational resources. For instance, Stable Video Diffusion is capable of generating a maximum of 25 frames, while AnimateDiff is limited to synthesizing sequences of up to 32 frames. To augment the capacity of these models to produce extended videos, we propose enhancements to the temporal modules within these models.

Firstly, the inherent functionality of 3D convolution layers to adaptively accommodate various scales has been previously validated through empirical studies \cite{jimenez2023mixture, bar2023multidiffusion, he2023scalecrafter}, even without necessitating fine-tuning. Consequently, we opt to retain the 3D convolution layers in their original form to preserve these capabilities. Secondly, regarding the temporal attention modules, research on large language models has demonstrated the potential for scaling existing models to accommodate longer contextual sequences \cite{xiong2023effective, chen2023longlora}. Inspired by these findings, we fine-tune the parameters within the temporal attention layers during the training process to enhance their efficacy over extended frame sequences. Thirdly, for the positional embedding layers, either static or trainable embeddings cannot be directly applied to longer videos. To circumvent this pitfall while ensuring compatibility with a wide array of existing video models, we use extended trainable parameters to replace the original positional embeddings. These extended trainable positional embeddings are initialized in a cyclic pattern, drawing upon the configurations of the pre-existing embeddings. Further, drawing inspiration from various adapter models \cite{hu2021lora, zhang2023adding}, we incorporate an additional identity 3D convolution layer subsequent to the positional embedding layer, aimed at learning long-term information. The central unit of this 3D convolution kernel is initialized as an identity matrix, and the remaining parameters are initialized to zero. The identity 3D convolution layer ensures that, before training, there is no alteration to the video representation, thereby maintaining consistency with the original computational process.

We apply our devised extending approach to Stable Video Diffusion \cite{blattmann2023stable}, which is a popular model within open-source communities for video synthesis. The comparative architectures, both pre and post-extension, are illustrated in Figure \ref{figure:architecture}. Because of the fundamental similarities that underpin the construction of temporal blocks within video synthesis models, our extending approach can also be applied to various video synthesis models.

\subsection{Post-Tuning}


\begin{figure}
  \includegraphics[width=1.0\linewidth]{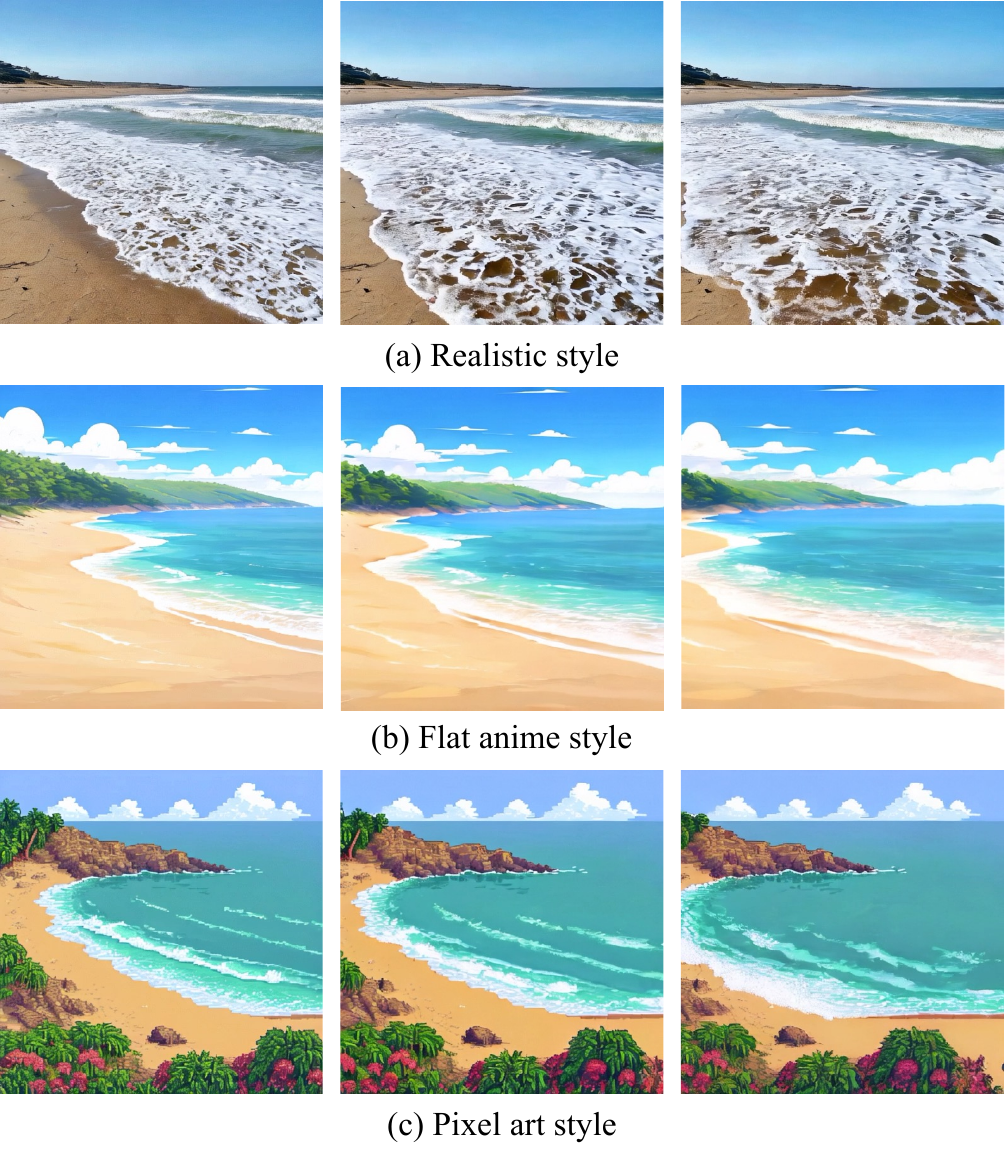}
  \caption{Examples in different styles generated by our Extended Stable Video Diffusion, where the first frame is generated by Stale Diffusion 3. The prompt is ``A beautiful coastal beach in spring, waves lapping on sand'', followed by the description of style.}
  \label{figure:style}
\end{figure}

After extending the temporal blocks in the video synthesis models, we enhance the model's abilities to generate extended videos via post-tuning. To circumvent potential copyright concerns with video content, we employed a publicly available dataset OpenSoraPlan\footnote{\url{https://huggingface.co/datasets/LanguageBind/Open-Sora-Plan-v1.0.0}}, which comprises 40,258 videos. These videos were sourced from copyright-free platforms, including Mixkit\footnote{\url{https://mixkit.co/}}, Pexels\footnote{\url{https://www.pexels.com/}}, and Pixabay\footnote{\url{https://pixabay.com/}}. The videos in this dataset maintain a resolution of $512\times 512$, inconsistent with the original resolution supported by Stable Video Diffusion. Given the model's design to accommodate varying resolutions, we opt to conduct the training at this resolution. ExVideo expands this capacity to 128 frames. Over such extended sequences, full training is deemed impractical because of the substantial computational requirements. Instead, we employed several engineering optimizations aimed at optimizing GPU memory usage. These optimizations are crucial for managing the increased computational load and facilitating efficient training within limited hardware resources:
\begin{itemize}
    \item \textbf{Parameter freezing}: All parameters except the temporal blocks are frozen.
    \item \textbf{Mixed precision training} \cite{micikevicius2017mixed}: We deploy a mixed precision training program by converting a subset of the model's parameters to 16-bit floating-point format.
    \item \textbf{Gradient checkpointing} \cite{feng2021optimal}: Gradient checkpointing is enabled in the model. By storing intermediate states during forward passes and recomputing gradients on-demand during the backward pass, this technique effectively decreases memory usage.
    \item \textbf{Flash Attention} \cite{dao2023flashattention}: We integrate Flash Attention to enhance the computational efficiency of attention mechanisms.
    \item \textbf{Shard optimizer states and gradients}: We leverage DeepSpeed \cite{rasley2020deepspeed}, a library optimized for distributed training, to enable shard optimizer states and gradients across multiple GPUs.
\end{itemize}
The loss function and the noise scheduler are consistent with the original model. The learning rate is $10^{-5}$ and the batch size on each GPU is 1. The training was conducted using only 8 NVIDIA A100 GPUs over one week. In order to ensure the stability of the training process, exponential moving averages are employed for the update of weights.

\section{Case Studies}

\subsection{Text-to-Video Synthesis}

By integrating existing text-to-image models, we can easily develop integrated pipelines capable of converting textual descriptions into videos. In this pipeline, the outputs from a text-to-image model are utilized as the foundational frames for the subsequent image-to-video model. Illustrative examples are shown in Figure \ref{figure:style}, wherein the foundational frames are synthesized by Stable Diffusion 3. The prompts direct the text-to-image model to create images across various styles. Our model subsequently generates fluent motion transitions from these high-quality images, even if styles such as flat anime and pixel art are not included in the training dataset. The Extended Stable Video Diffusion model preserves and extends the generalization capabilities in the original model.



\subsection{Visualization of Training Process}

\begin{figure}
  \includegraphics[width=1.0\linewidth]{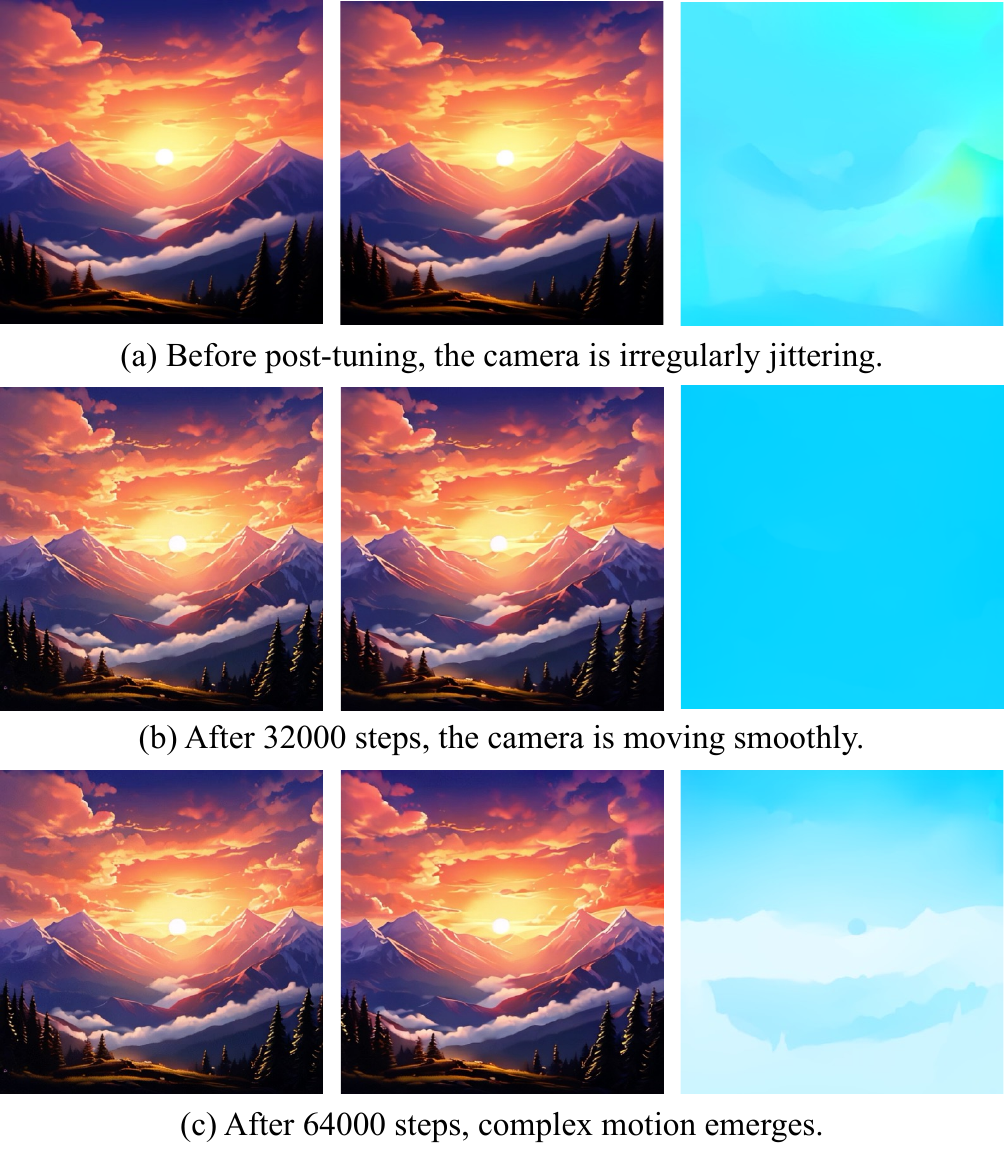}
  \caption{Video examples in different training phases. The first frame is generated by Hunyuan DiT, and the prompt is ``sunset, mountains, clouds''. We present the optical flow to visualize the motion, where pixels with similar colors are moving in similar directions.}
  \label{figure:visualization}
\end{figure}

We investigated the evolution of the model's capabilities during the training process. Figure \ref{figure:visualization} presents the generated videos that exemplify the model's performance at three distinct phases of training. It is difficult to present the dynamics using still images, thus we present the optical flow, computed by RAFT \cite{teed2020raft}, to the right of each example for a clearer demonstration of motion. Initially, before training, the extended model architecture was solely capable of guaranteeing the structural integrity of the video frames, which suffered from pronounced jittering artifacts. Progressing through the training, after 32,000 steps, the model began to produce videos displaying smooth camera movements. With continued training up to 64,000 steps, the model further advanced to create complex motions, such as clouds and mountains moving with nuanced, layered speed. The model effectively understands the depth and spatial relationships within the scene. This example intuitively illustrates the process of the model learning long-term information.

\subsection{Adaptive Resolution}

\begin{figure}
  \includegraphics[width=1.0\linewidth]{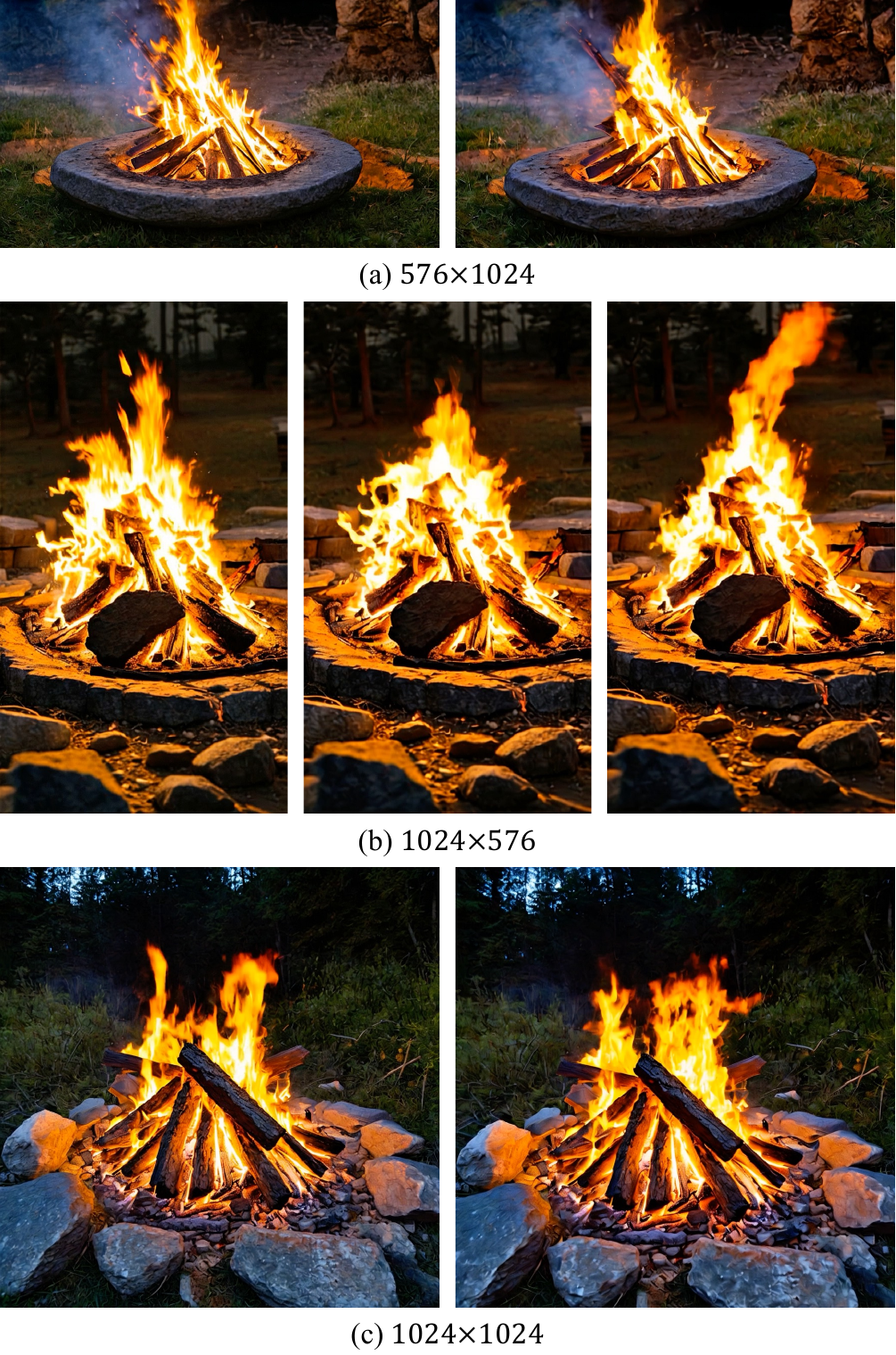}
  \caption{Video examples in various resolutions. The first frame is generated by Stable Diffusion 3, and the prompt is ``bonfire, on the stone''.}
  \label{figure:resolution}
\end{figure}

We also ascertain the performance of the extended Stable Video Diffusion across various resolutions. Several video examples are illustrated in Figure \ref{figure:resolution}. These examples demonstrate that the model, with common aspect ratios, adeptly generates videos in higher resolutions. This capability not only highlights the robustness and generalizability of the base model but also underscores the effectiveness of our post-tuning methodologies in enhancing model performance.

\subsection{Comparison with Other Models}

\begin{figure*}
  \includegraphics[width=1.0\linewidth]{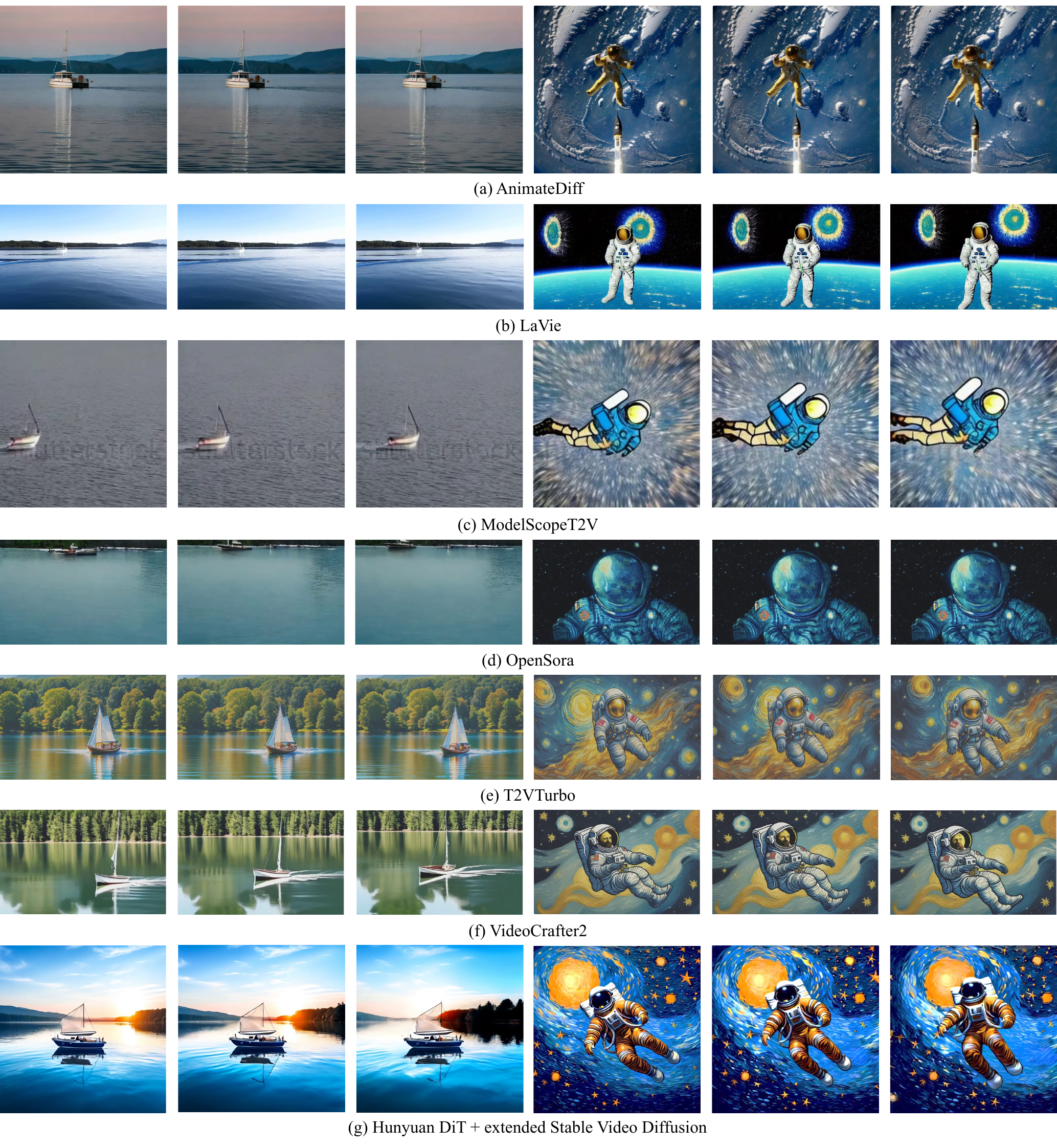}
  \caption{Visual comparisons of text-to-video results from several existing video synthesis models and our Extended model. The prompts are ``a boat sailing smoothly on a calm lake'' and ``an astronaut flying in space, Van Gogh style''. In our pipeline, the first frame is generated by Hunyuan DiT, and our extended Stable Video Diffusion generates the video according to the first frame.}
  \label{figure:compare}
\end{figure*}

To evaluate the performance of our model, we conducted comparative analyses against several existing video synthesis models. Illustrative outcomes from these models, including AnimateDiff \cite{guo2023animatediff}, LaVie \cite{wang2023lavie}, ModelScopeT2V \cite{wang2023modelscope}, OpenSora\footnote{\url{https://github.com/hpcaitech/Open-Sora}}, T2VTurbo \cite{li2024t2v}, VideoCrafter2 \cite{chen2024videocrafter2}, alongside our extended Stable Video Diffusion, are displayed in Figure \ref{figure:compare}. The videos generated by the baseline models are collected from GenAI-Arena\footnote{\url{https://github.com/ChromAIca/VideoGenMuseum}} \cite{ku2023imagenhub}. The text-to-image model utilized in our pipeline is Hunyuan DiT \cite{li2024hunyuan}. A critical observation from this comparison is that the majority of existing video synthesis models usually generate videos with minimal motion dynamics. In contrast, owing to the post-tuning processes applied over extensive temporal duration, our extended model demonstrates a superior capability to generate videos with significant movements. This differential outcome underscores the advanced generative performance of our model.

\section{Limitations}

While ExVideo can enhance the capabilities of video diffusion models, the post-tuned version continues to be constrained by the inherent limitations of its foundational model. Notably, the extended Stable Video Diffusion struggles to accurately synthesize human portraits, leading to frequent instances of truncated frames. To develop a model capable of synthesizing high-quality long videos, it is imperative to train a robust base model. Nevertheless, due to limitations in resources, we are unable to independently pre-train a large video synthesis model. Consequently, we eagerly anticipate the release of open-source models in the future to advance our research endeavors.

\section{Conclusions and Future Work}

In this paper, we delve into the enhancement of video diffusion models through post-tuning. Specifically, we propose a post-tuning approach called ExVideo, which can extend the duration of generated videos and release the potential of video synthesis models. Based on Stable Video Diffusion, our approach achieves a quintupling in the number of frames, while preserving the original generalization abilities. ExVideo is designed within the constraints of limited computational resources, thus it is exceptionally memory-efficient. By integrating this method with other open-source technologies, we facilitate pipelines conducive to the production of high-quality videos. However, the post-tuned models still face the limitations of the base model. To further improve the performance, we will try to train the model on larger datasets in the future.

\begin{acks}
This work was supported by the National Natural Science Foundation of China under grant Number 62202170, Fundamental Research Funds for the Central Universities under grant Number YBNLTS2023-014, and Alibaba Group through the Alibaba Innovation Research Program.
\end{acks}

\bibliographystyle{ACM-Reference-Format}
\bibliography{sample-base}

\end{document}